\documentclass[conference]{IEEEtran}
\IEEEoverridecommandlockouts
\usepackage{cite}
\usepackage[normalem]{ulem}
\usepackage{amsmath,amssymb,amsfonts}
\usepackage{algorithmic}
\usepackage{graphicx}
\usepackage{textcomp}
\usepackage{xcolor}
\usepackage{url}
\usepackage{caption}
\usepackage[shortlabels]{enumitem}

\usepackage{subcaption}
\def\BibTeX{{\rm B\kern-.05em{\sc i\kern-.025em b}\kern-.08em
    T\kern-.1667em\lower.7ex\hbox{E}\kern-.125emX}}
\makeatletter 
\newcommand{\linebreakand}{%
  \end{@IEEEauthorhalign}
  \hfill\mbox{}\par
  \mbox{}\hfill\begin{@IEEEauthorhalign}
}
\makeatother 

\setlist[description]{
  leftmargin=1em, itemindent=-.5em,
  itemsep=-0.25em, topsep=0em
}

\begin{document}

\title{iDML: Incentivized Decentralized Machine Learning}

\author{
    \IEEEauthorblockN{Haoxiang Yu\IEEEauthorrefmark{1}, Hsiao-Yuan Chen\IEEEauthorrefmark{1}, Sangsu Lee\IEEEauthorrefmark{1}, Sriram Vishwanath\IEEEauthorrefmark{1}, Xi Zheng\IEEEauthorrefmark{2}, Christine Julien\IEEEauthorrefmark{1}}
    \IEEEauthorblockA{\IEEEauthorrefmark{1}Department of Electrical and Computer Engineering, University of Texas at Austin
    \\\{hxyu, littlecircle0730, sethlee, sriram, c.julien\}@utexas.edu}
    \IEEEauthorblockA{\IEEEauthorrefmark{2}Department of Computing, Macquarie University, james.zheng@mq.edu.au}
}

\maketitle

\begin{abstract}
With the rising emergence of decentralized and opportunistic
approaches to machine learning, end devices are increasingly tasked with training deep learning models on-devices using crowd-sourced data that they collect themselves. These approaches are desirable from a resource consumption perspective and also from a privacy preservation perspective. When the devices benefit directly from the trained models, the incentives are implicit – contributing devices' resources are incentivized by the availability of the higher-accuracy model that results from collaboration. However, explicit incentive mechanisms must be provided when end-user devices are asked to contribute their resources (e.g., computation, communication, and data) to a task performed primarily for the benefit of others, e.g., training a model for a task that a neighbor device needs but the device owner is uninterested in. In this project, we propose a novel blockchain-based incentive mechanism for completely decentralized and opportunistic learning architectures. We leverage a {\em smart contract} not only for providing explicit incentives to end devices to participate in decentralized learning but also to create a fully decentralized mechanism to inspect and reflect on the behavior of the learning architecture. 
\end{abstract}


\section{Introduction}
\label{introduction}
With the continued development of machine learning techniques and the increasing computing power of personal devices, machine learning has been heavily applied to various user-facing tasks, such as photo classification, input prediction, smart home systems, etc \cite{wide_use_ml}. These tasks commonly rely on deep neural networks (DNNs) that learn patterns over large data sets. Achieving high accuracy for those tasks requires a massive amount of training data. However, largely because of privacy concerns and limited communication bandwidth between end-user devices and the cloud, gathering sufficient data to train robust models is challenging~\cite{9369019}.

To support increased user privacy and enable applications to operate in communication-constrained environments, machine learning has expanded to include a variety of distributed learning approaches, where end devices assume some of the burden associated with training. Most well-known are approaches to {\em federated learning}, where a central server coordinates a vast array of end-user devices to iteratively train a shared global model. End devices receive a snapshot of a model, perform some rounds of training of the model using locally available data, then send model updates to the central server. The server aggregates the received updates, generates a new global model, then repeats the process. In a pure federated learning design, the central server does not have any training data of its own, which  makes it challenging for it to detect whether end nodes' contributions to the shared model are valid. 

In some contexts, relying on a central server to coordinate the learning process of a distributed set of end devices is either not possible (e.g., the devices are not well connected to the Internet) or not optimal (e.g., the devices desire to learn more personalized models rather than a single global model). Thus, approaches have emerged that completely {\em decentralize} the learning process and instead rely only on {\em opportunistic} device-to-device collaboration among end-user devices. Existing research has shown that such decentralized approaches can achieve high model accuracy under a variety of conditions~\cite{9439130,colin2016gossip}. Throughout this paper, we refer to such approaches that rely only on device-to-device encounters to train models on-device as {\em decentralized opportunistic learning}. Decentralized opportunistic learning uses ephemeral encounters to support on-device training in collaboration with encountered neighbors and their local (private) data. As a straightforward example, gossip learning~\cite{colin2016gossip} simply decentralizes the entire learning process, using only opportunistic pair-wise exchanges between nodes to incrementally learn the same shared global model learned in a centralized approach. As a second example, in {\em Opportunistic Collaborative Learning} (OppCL)~\cite{9439130}, rather than exchanging gradients learned while seeking convergence to a global model, devices learn individual models influenced by the unique combination of a device's own data and the data it encounters. In OppCL, when a device encounters another device, it shares its personal model parameters with the neighbor who then uses their own local data to compute an update specific to the learning device's goal.

Decentralized opportunistic learning requires many activities of end devices, and each of these activities requires devices to contribute resources. By its nature, any approach to decentralized learning relies on end devices to donate their data, computational, and communication resources. As such, {\em incentivizing participation} in the learning process remains a significant open challenge to realizing decentralized opportunistic learning in practice. To support on-device training, devices must collect training data, which may need to be labeled with valid labels. For instance, if the application is to train an object  detection system~\cite{zhao2019object}, the data collected are photographs, and each photo must be labeled with the primary object in the frame. If the application is next word prediction~\cite{fedlearning_wordprediction}, the data collected is the potentially sensitive text that the user types, with the labels coming ``for free'' based on the sequences of words input. Data collection and labeling consume resources either explicitly (e.g., the user actively labels data) or implicitly based on the computation required to label. The data must be stored, and it must be used to train, which requires computational resources from each end node. And as the participating devices collaborate, their exchanges also consume valuable communication resources.

In addition, while the design of decentralized opportunistic learning protects a user's privacy by preventing the direct sharing of raw data, it is difficult for the recipient of a model update to ascertain whether the other parties participating are faithfully holding up their obligations to the algorithm. Further, because the entire process is decentralized and opportunistic, it is difficult to inspect and reflect on the provenance of the model –-- to do so requires storing and sharing what happens during training, how the model's accuracy evolves, what data is used to train the model, etc. More generally, due to the decentralized architecture, it can be difficult to avoid or detect attackers of all kinds, including those seeking to interrupt training and those seeking to subvert it.

The completely decentralized and distributed nature of this problem naturally lends itself to blockchain~\cite{zheng2018blockchain}, a type of distributed ledger technology, and associated smart contracts, a term originally referring to the automation of legal contracts~\cite{szabo1997formalizing} that now popularly refers to the code scripts that run synchronously on multiple nodes of a distributed ledger~\cite{8847638}. The natural mechanisms for blockchain to store and compute in a decentralized way makes it ripe for integration with decentralized opportunistic learning.
In particular, we propose to combine the advantages of decentralized opportunistic learning with the smart contract to (1)~robustly incentivize participation in opportunistic learning; (2)~provide a means to record and reflect on model provenance in decentralized learning; and (3)~examine a new opportunistic decentralized learning pipeline that is robust to a well-defined attack model.

This paper makes the following concrete contributions:
\begin{itemize}
    \item We developed a blockchain-based {\em smart contract} that can be applied to decentralized opportunistic learning structures that incentivize participants to contribute to the learning process.
    \item We design mechanisms into the smart contract to support using it to record and examine essential information related to the progress of a completely decentralized learning execution.
    \item We demonstrate how the use of this smart contract can prevent attacks on both the new incentive mechanism and the decentralized opportunistic learning algorithm.
    \item We evaluate the incentive system in simulation with different opportunistic yet fully decentralized learning algorithms in place. The developed techniques will be released as open-source artifacts.
\end{itemize}

The paper is organized as follows: in Section \ref{related_works}, we discuss the related work in decentralized learning, smart contracts, and incentive mechanisms. In Section \ref{system}, we present our system design and smart contract workflow. In Section \ref{evaluation}, we use a simulation system to benchmark the behavior of our proposed system with two different learning algorithms, two different datasets, and two typical data distributions in decentralized learning (IID/Non-IID). In the final section of the paper, we conclude and explore possible future work.
\section{Related Work}
\label{related_works}
In recent years, the research community has introduced several new models of distributed learning, which provides new opportunities to benefit from users’ data and comptuational resources with a minimum risk of privacy leaks. The related work of this research has four major components: (1)~distributed and decentralized learning approaches; (2)~blockchain and smart contracts; (3)~incentive mechanisms for distributed and decentralized learning; and (4)~attacks on incentive mechanisms and decentralized learning algorithms.

\subsection{Distributed Learning}

The growth in mobile device capabilities and the desire for data privacy have motivated federated and decentralized learning. In federated learning, devices collaborate to construct a global model in a way that maintains the data privacy for each individual device~\cite{kairouz2021advances,fedavg}. Applications of federated learning in pervasive computing include wake word detection, also known as keyword spotting in smart home voice assistants, where federated learning protects the potentially private audio data collected at users' end devices~\cite{leroy2019federated}.
Another classic federated learning application
is next word prediction on a mobile device keyboard~\cite{hard2018federated}. 
Other applications have also explored
on-device image classification and image processing~\cite{xiong2019antnets}.

These traditional approaches to federated learning assume the existence of a single central coordinator, and the goal of the devices in 
model aggregation
is to
collaboratively
learn a single global model that is shared among all of the participants. This differs from decentralized and opportunistic approaches, in which the learning process progresses only through opportunistic encounters between devices. In this work, we consider two different styles of decentralized opportunistic learning: gossip-based learning~\cite{colin2016gossip} and opportunistic collaborative learning~\cite{9439130}. In gossip-based learning, each time a device encounters another, it performs a {\em merge-update-send} cycle. In this process, the encountered neighbor shares its locally trained model, the receiving device merges the received model into its local model, then the receiving device performs training on the merged model using only the local data. Of course, this is commonly a two-way process; in a bi-directional encounter, both devices perform the {\em merge-update-send} process independently. In gossip-based learning, the goal is for an intermittently interconnected set of nodes to eventually learn a consistent shared global model. In contrast, in
the opportunistic collaborative architecture, each device seeks to learn its own personalized model but to opportunistically integrate learning based on neighbors' local data on-demand. In opportunistic collaborative learning, a device that encounters another device engages in a {\em send-train-return-merge} cycle. In this process, when a device encounters a neighbor, it sends the neighbor a copy of the locally stored model. The neighbor performs a round of training on that model using the neighbor's local data then returns the updated model. The device merges the updated model with the stored model and continues training with the local data.

As can be seen from these descriptions, both models ask devices to contribute local resources to support the learning process for themselves and other devices. A significant challenge in designing such a decentralized and opportunistic system is incentivizing devices to contribute their local resources for the greater good and in a secure manner. This is the main problem we tackle in this paper.

\subsection{Blockchain and Smart Contract}

In this paper, we use a smart contract as the program to execute our approaches with fairness, security, and auditability. In this context, a smart contract is a computer program that can automatically execute on a blockchain platform. In particular, we leverage blockchain support for executing smart contracts in order to implement the incentive mechanism described in this paper. Once a smart contract has been deployed, all of the participants can review whether the contract is fair and whether executing it is in their best interest. In addition, all of the actions made by participants also can be viewed by other participants within the smart contract; existing research also shows that executing a smart contract on the blockchain can guarantee the correctness of the voting result~\cite{8887296}.

While blockchain is commonly associated with digital currency, several platforms have emerged that apply blockchain beyond digital currency. In particular, the Ethereum ecosystem allows one to run user-defined programs on the blockchain~\cite{wood2014ethereum}. Using Ethereum, for instance, a user can translate their contract into a programming language, compile the program to bytecode, and execute it on the Ethereum Virtual Machine (EVM). Because the contract in the EVM can execute by itself, we call the
program a smart contract.

\subsection{Incentive Mechanisms}
Our goal in this work is to provide digital incentives to encourage users to contribute their resources to decentralized learning algorithms. Existing work has applied game theory to design incentive mechanisms for centralized federated learning~\cite{9094030,9006179,9006327,9780573}. In game theory, participants make individual decisions that affect other participants' actions. 
Zhang et al. used a Stackelberg game to incentivize participation in federated learning by using the server as the leader and the edge nodes as the
followers \cite{9184072}. The system works in two stages: first, the server announces the total reward then the edge nodes use that information to determine their training strategies to maximize their individual rewards. 
Lim et al. used coalitional game theory to hierarchically incentivize federated learning by rewarding model owners based on their marginal contributions to the
model aggregation
process \cite{9057543}. 

Researchers have also considered auctions to support incentives in federated learning setting. Prior work has explored several different auction strategies, including sealed-bid, forward, reverse, double, and combinatorial auctions~\cite{9223754,9094030,9488743}. Further, Zhu et al. explored different aspects of the auction process, including power, sensing, data resources, and training resources as they relate to federated learning~\cite{9780573}. However, game theory and auction approaches are limited because they assume that all of the participants act in the same way. Also, these approaches are not safe and lack of auditability. For instance, they cannot detect attacks nor record essential information for the learning progress.

Beyond game theoretic and auction-based mechanisms, recent work has also explored the application of blockchain to design incentive mechanisms, in particular for federated learning approaches that involve a central coordinator. Wang et al. designed a federated learning framework that allows clients to share local gradients as transactions in the blockchain. Designated workers validate these transactions and pack them into blocks \cite{9006179}. Similarly, FLChain, proposed by Bao et al., is a blockchain-based ecosystem that ensures provenance and maintains auditable aspects of the federated learning model \cite{8905038}. These works show us how to effectively find and remove dishonest nodes. They have also shown us the possibility to maintain a user’s privacy while also making the learning process auditable. More generally, they have characterized the goals of confidentiality, auditability, and fairness \cite{8905038}, which we will also use to frame our research.

However, these approaches to applying blockchain to incentives in distributed learning focus exclusively on federated learning settings that involve a central coordinator, and these approaches are not directly applicable to a completely decentralized learning structure where we do not assume that all of the devices request the global model to conduct the next round of training. Further, their approaches to information sharing to support the learning process itself rely heavily on the blockchain, which in turn incurs very high costs and limits the potential applicability of the approach to applications that can justify this cost. In contrast, we propose to interact with the blockchain via smart contract for only essential interactions related to model provenance and incentives. In doing so, we keep the distributed learning protocol itself off of the blockchain, allowing a more agile and resource-conscious implementation.

\subsection{Attackers in Incentivized Decentralized Learning Settings}

In general, the decentralized learning and the incentive mechanism proposed by this paper faced two major types of attack, which we refer to as {\em incentive mechanism attacks} and {\em learning attacks}. Although the focus of this paper is on providing incentives for the participants, the design can also protect the system from both types of attacks. 

{\bf Incentive Mechanism Attacks.} An {\em incentive mechanism attack} is an attack that is enabled by the design of the incentive mechanism itself. In such an attack, due to the motivation of the rewards provided by the incentive mechanism, an attacker will try to get a reward without performing the real work (e.g., by providing fake or useless training information). To thwart incentive mechanism attacks, we carefully design our incentive mechanism to both detect and prevent these, as described in Section~\ref{system} below.

{\bf Learning attacks.} The distributed nature of decentralized learning opens itself to a variety of attacks that can disrupt the process. In particular, these attacks can be categorized as: (1)~evasion attacks, in which an attacker aims to cause the model to make incorrect predictions through the use of adversarial testing data~\cite{evasion_attack}; (2) poisoning attacks, in which the attacker contaminates the training data to cause the model to misclassify the input~\cite{shafahi2018poison}; and (3) exploratory attacks in which an attacker reverses the model in an attempt to recover information about the training data~\cite{fi13030073}. In the context of the decentralized opportunistic learning approaches addressed in this paper, the poisoning attack is the most relevant; in the remainder of this paper, we refer to such attacks simply as {\em learning attacks}.

There are some existing solutions to detect these learning attacks, but to date, they have only been designed for use in centralized federated learning settings. For instance, 
Sun et al., identifies the {\em backdoor attack} on federated learning and propose a norm thresholding method, in which the central server rejects any model updating having a norm larger than some threshold $M$ \cite{really_backdoor}. They show that the performance depends on the proportion of nodes in the system that are adversaries Krum, proposed by Blanchard et al., is an aggregation rule that can be used to make a distributed stochastic gradient descent conmputation resilient to Byzantine failures~\cite{blanchard2017machine}.

Since in the decentralized setting, clients have full control of the data as they are the only owners of the data, a simple learning attack can also be launched by simply flipping the labels of training samples from the target class \cite{shafahi2018poison} before performing the training. 

Existing work on addressing these attacks has primarily viewed these attacks more at the node or model level rather than at the level of the decentralized system. Moreover, existing solutions rely on a central server's global view i.e., they support centralized federated learning and not decentralized approaches. In contrast, by using smart contracts and voting mechanisms, our approach allows any participant to identify and drop models that do not improve the accuracy of the model provided by the other participants. At the system level, users who produce harmful or unimproved contributions over time will be punished and automatically eliminated from the environment over time.
\section{System Design}
\label{system}
\begin{figure}[!t]
  \centering
  \includegraphics[width=.48\textwidth]{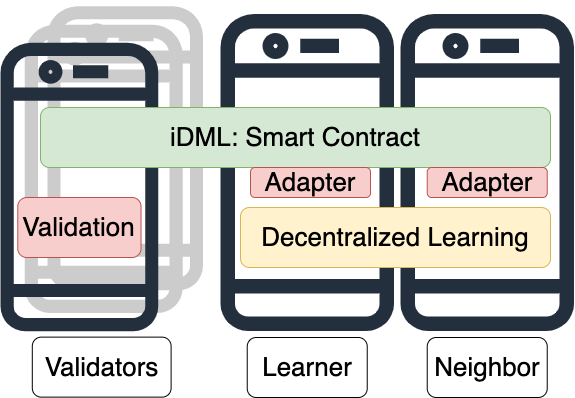}
  \caption{System Architecture}
  \label{fig:system_architecture} 
\end{figure}

This work provides a generic platform to incentivize participation in decentralized opportunistic learning. 
In decentralized opportunistic learning, this participation, which includes contributing resources like computation, communication, and data, is essential to the success of the learning process, in particular with respect to the achievable accuracy of a learned model.
Fig.~\ref{fig:system_architecture}
shows, abstractly, how this incentive mechanism, called iDML for {\underline{i}ncentivied \underline{D}ecentralized \underline{M}achine \underline{L}earning}, relates to existing decentralized and opportunistic learning algorithm implementations.

In iDML, there are three roles: (1) the \textbf{Learner} is the participating device that seeks to benefit from a particular encounter in a decentralized opportunistic learning algorithm; (2) the \textbf{Neighbor} is the participating device that contributes their resources so the learner can learn; and (3) the \textbf{Validators} are the participants who validate the neighbor's contribution. A device can take on more than one role at different times; for a given encounter, the validators must be distinct from the learner and neighbor, however, in a symmetric protocol (e.g., gossip learning), two nodes in an encounter may simultaneously benefit from each other. In this case, iDML treats the encounter as two encounters, one with each device in the learner role and the encountered device in the neighbor role.

As Fig.~\ref{fig:system_architecture} shows, we design iDML as a plug-in; the smart contract implementation is kept separate from the core decentralized learning functionality, making it portable and generalizable. In order for iDML to interoperate with an existing decentralized learning algorithm, we implement an algorithm-specific {\em adaptor}, whose role is effectively to choreograph the interactions between iDML above and the implementation of decentralized learning, below. In addition, we also implement algorithm-specific validators. As described in more detail below, the specific steps of validation vary based on the learning algorithm in use. Above, iDML is responsible for maintaining the smart contract and coordinating across the participants in the three roles.

\begin{figure}[ht]
  \centering
  \includegraphics[width=.48\textwidth]{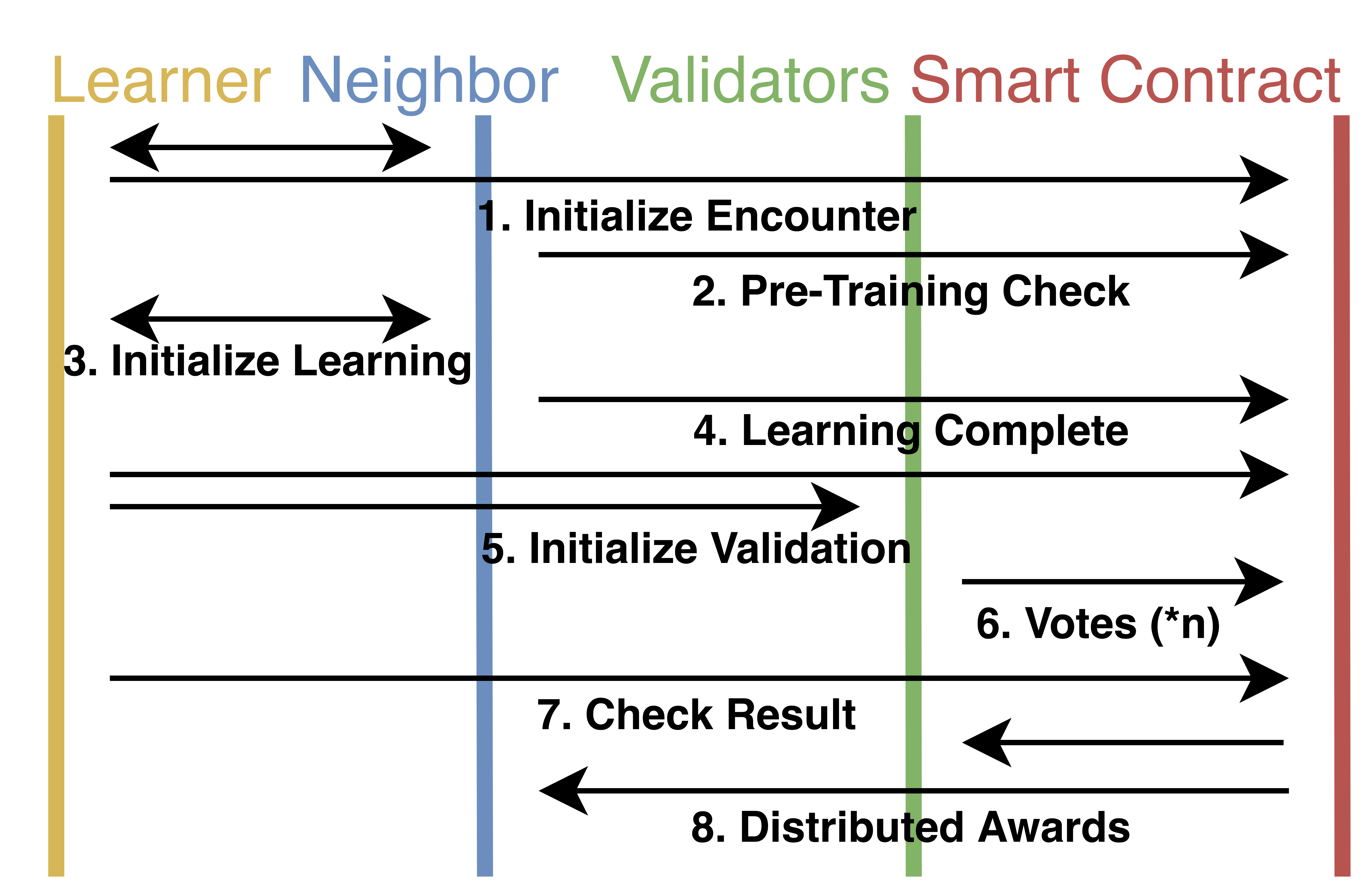}
  \caption{A system view of iDML. Except for step 1 and 3, all other steps belong to iDML.}
  \label{fig:system_layer} 
\end{figure}

\subsection{Motivating Participation in Decentralized  Learning}
The primary goal of our system is to motivate the participants to actively participate in the decentralized opportunistic learning process and to provide compensation for their contributed resources. The overall flow of the iDML approach is shown in Fig.~\ref{fig:system_layer}. While the figure depicts many steps, the process is, overall, intuitive, and entails nine steps:
\begin{enumerate}
    \item The learner and neighbor encounter one another, and the underlying decentralized learning protocol is triggered.
    \item Before learning starts, the iDML Adapter is invoked, and the learner is requested to pay a predefined and agreed-upon amount of tokens into the smart contract. The neighbor device checks this prepayment before re-commencing the learning activity. 
    \item Once the prepayment is done and verified, the learner and neighbor complete whatever collaborative task is demanded by the decentralized learning algorithm.
    \item After the collaborative task completes, both the learner and neighbor inform the smart contract.
    \item Once the validators receive a validation request, they perform the validation by using the algorithm-specific validation implementation. 
    \item After the validators validate the model, they vote by sending the result to the smart contract.
    \item After a predefined number of votes has been received on the smart contract or a max time has passed, any one party requests the result within the smart contract.
    \item Then, the smart contract distributes the awards or applies the penalty to the neighbor and validators based on the predefined and consensus methods. 
\end{enumerate}

We next present the workflow of our approach as centered around the smart contract. Throughout, we connect each step back to the numbered steps in Fig.~\ref{fig:system_layer}, though we omit step 3, which focus on the decentralized learning process and not on the role of the smart contract.

\begin{figure}[!t]
  \centering
  \includegraphics[width=.95\columnwidth]{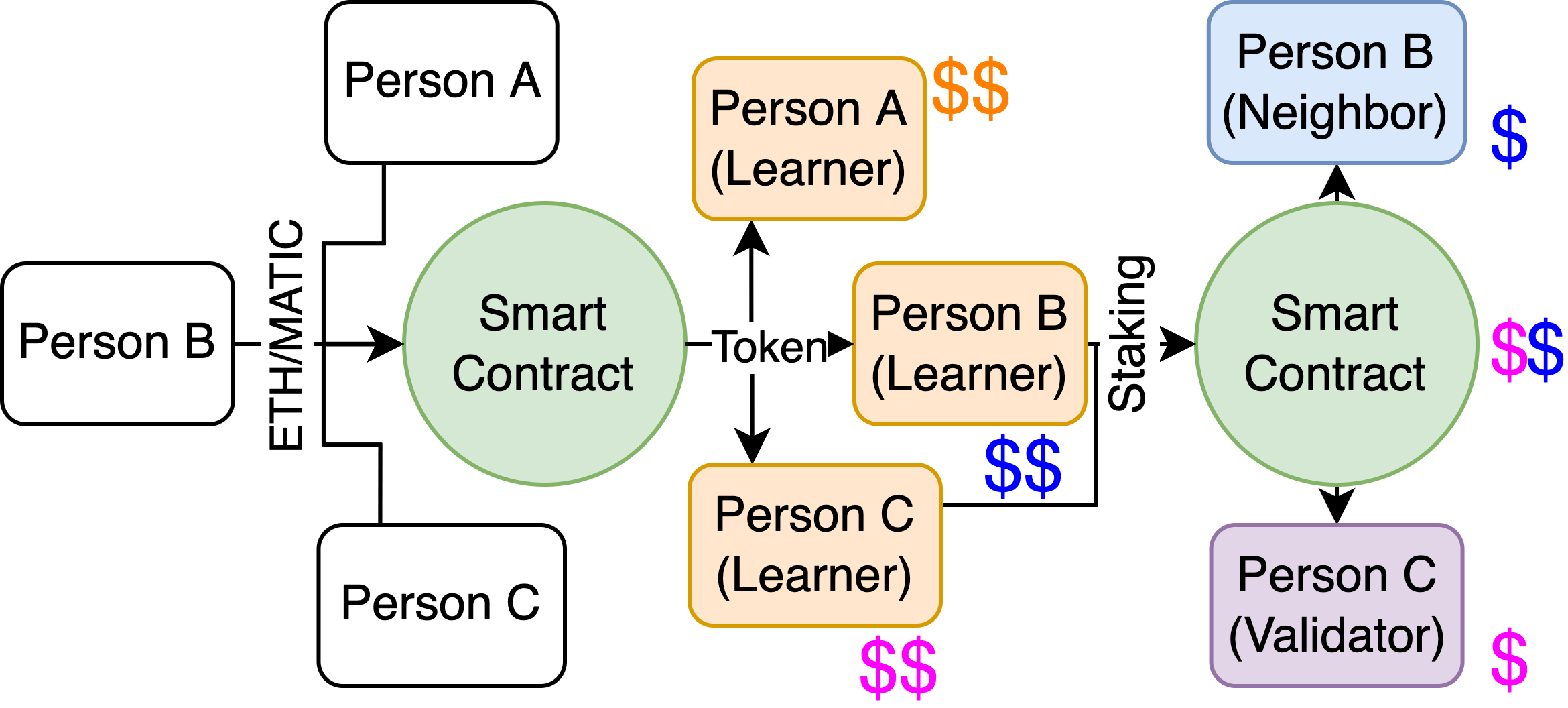}
  \caption{Initializing the smart contract ({\bf Step 0}).}
  \label{fig:step0} 
\end{figure}
\textbf{Step 0 (Initialization)}: To initialize the use of the smart contract, prior to any of the steps shown in Fig.~\ref{fig:system_layer}, the individuals who own the devices participating in decentralized opportunistic learning must join the process by exchanging their digital currencies with tokens, a form of digital currency, supported by our smart contract with the ERC-20 token standard\footnote{https://ethereum.org/en/developers/docs/standards/tokens/erc-20/}. The smart contract has a predefined rate with one digital currency but it can use that digital currency as the bridge to exchange with other digital currencies. For a participant to serve as a {\bf Neighbor} or {\bf Validator} in iDML, they need to stake a predefined amount of tokens to smart contract as {\em pledges}.  The award is paid from the {\bf Learner} to the {\bf Neighbor} and {\bf Validator} once a specific opportunistic learning process is successfully finalized. The complete initialization process is shown in Fig.~\ref{fig:step0}.

\begin{figure}[!t]
  \centering
  \includegraphics[width=.85\columnwidth]{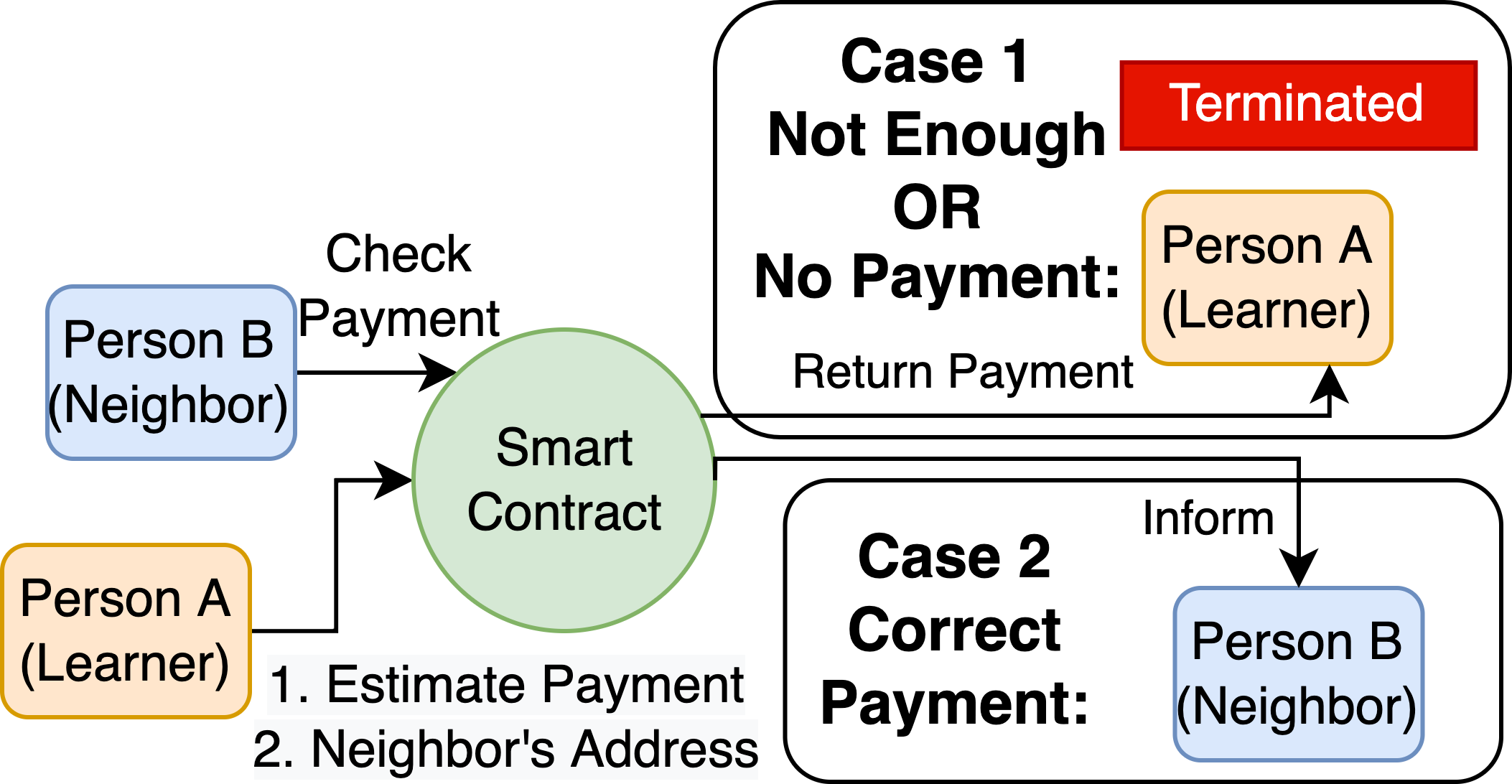}
  \caption{Pre-payment and checking pre-payment before initiating encounter-driven training process ({\bf Step 1 and 2}).}
  \label{fig:step2} 
\end{figure}

\textbf{Step 1 (Initialize Encounter) and Step 2 (Pre-Training Check):}  After an encounter is initiated and two participants tentatively agree to collaborate, the two participants (i.e., the {\bf Learner} who will benefit from the encounter and the {\bf Neighbor} who will provide resources to support the decentralized learning algorithm) both check in with the smart contract before continuing. In particular, the {\bf Learner} informs the smart contract of the identity of the neighbor it will collaborate with and authorizes the prepayment (second arrow in Fig. \ref{fig:system_layer}). The method of computing the right amount of payment is based on a consensus between the neighbor and the learner. The prepayment is based on the communication and computational resource requirements of the decentralized learning process and can be defined globally with all of the participants or privately negotiated during the encounter step. The {\bf Neighbor} will check the validity of the prepayment with the smart contract. If the prepayment is insufficient, the {\bf Neighbor} can reject the collaboration, and the prepayment will be returned to the {\bf Learner}'s account in the smart contract. Otherwise, the {\bf Learner} and {\bf Neighbor} will collaborate as defined by the decentralized learning algorithm. This pre-training check-in with the smart contract is depicted in Fig.~\ref{fig:step2}.

\begin{figure}[!t]
  \centering
  \includegraphics[width=.275\textwidth]{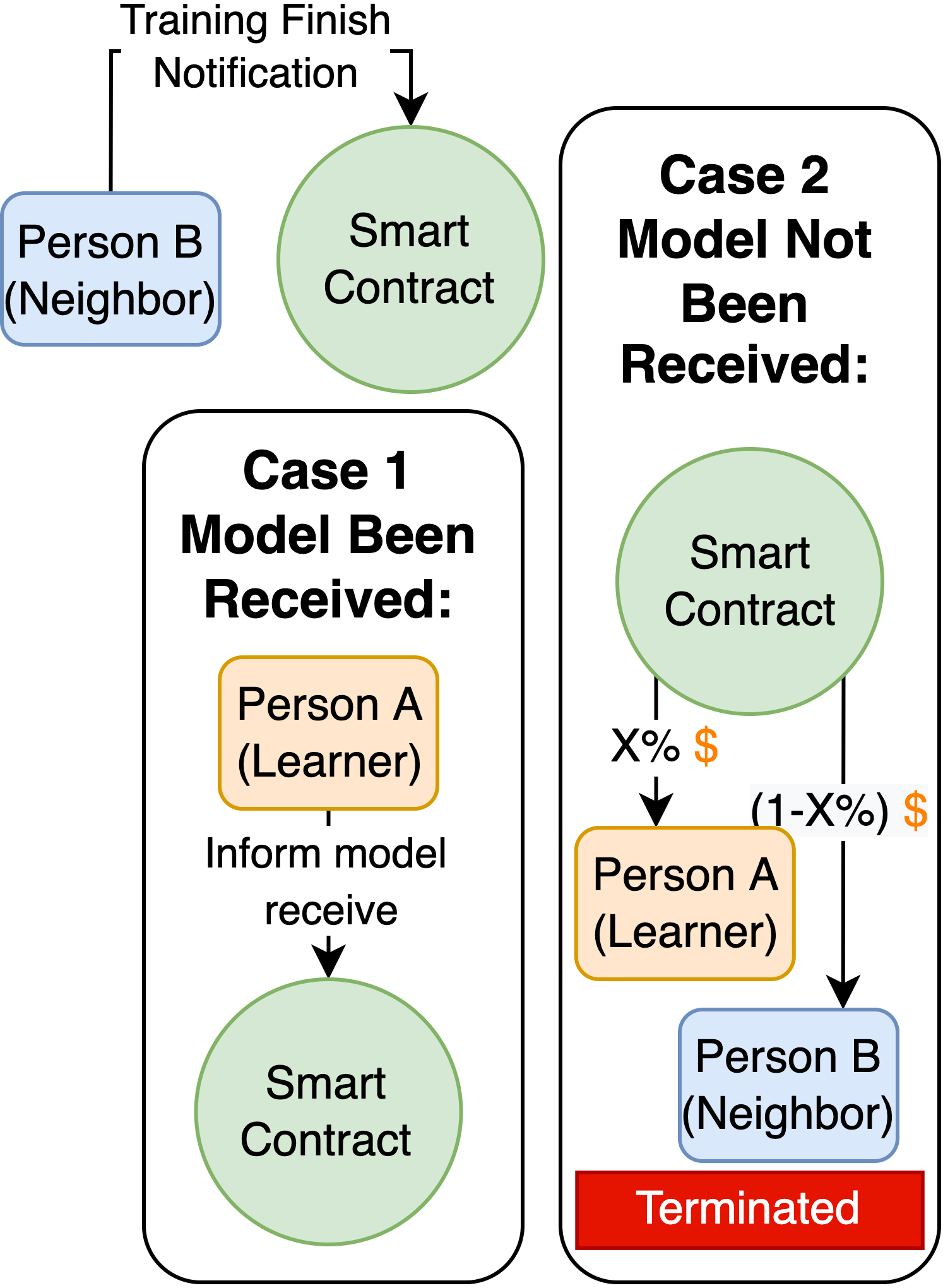}
  \caption{Completing the Learning Task ({\bf Step 4}).}
  \vspace{-.5cm}
  \label{fig:step5} 
\end{figure}

\textbf{Step 4 (Learning Complete):} Step 3 (initialize learning) in Fig.~\ref{fig:system_layer} completes by the decentralized opportunistic learning algorithm, without any interaction with the smart contract. However, when the collaboration process completes and
the {\bf Learner} is given the result (e.g., model parameters), the iDML Adapter layer on the {\bf Neighbor} intercepts and makes a copy of the result and sends the ``learning complete" signal and the result's md5  to the smart contract in iDML. In addition, the {\bf Learner} also reports to the smart contract the result's md5 through the iDML Adapter layer on the Learner's side. At this step, there are two possible outcomes. In the first case
(Case 1 in Fig.~\ref{fig:step5}), assuming the results from the {\bf Learner} and {\bf Neighbor} match, the validation process can begin. On the other hand
(Case 2 in Fig.~\ref{fig:step5}), several different possibilities may prevent the process from continuing. Due to the nature of decentralized learning, either the {\bf Learner} or the {\bf Neighbor} may not reachable or may be unable to send or receive the results of the learning process. In this case, as the {\bf Neighbor} has likely contributed its resources to the training process, a portion of the prepayment will be given to the {\bf Neighbor} as a show of good faith. On the other hand, as the {\bf Learner} did not benefit from the collaboration, the rest of the prepayment will return to the {\bf Learner}. 
        
In determining the magnitude of the small payment to a {\bf Learner}, we seek a balance that prevents the participation of a malicious learner and/or neighbor. Over time, if a particular {\bf Neighbor} or {\bf Learner} builds up a history of incomplete interactions within the smart contract, this information can be used in Step 2 before continuing with future encounters. For instance, the smart contract can check the history of a {\bf Neighbor}'s block by block in the underlying blockchain for deeper verification.

\textbf{Step 5 (Model Validation):} To kick off the validation step, the {\bf Learner}shares the learning inputs and results with the validators (first arrow for step 5) and notifies the smart contract that the validation has been initialized (second arrow for step 5). Then, the validators begin to verify whether the {\bf Neighbor}'s contribution is solid. The validators are selected based on the same criteria as the neighbor. On the smart contract, the learning result is indexed by using an md5 hash of the model weights and the validator's id, which prevents the learner from sending the wrong model to the validators. As shown in Fig.~\ref{fig:system_architecture}, the verification method is dependent on the particular decentralized learning algorithm used. This step invokes a number of validators suitable for the learning algorithm (e.g., empirically tuned). Each validator {\em votes} on whether they believe the result is valid, based on an assessment of an updated model relative to the {\bf Learner}'s previous model. That is, validator
checks whether the {\bf Neighbor}'s contribution to learning was positive. Because we allow for some uncertainty in the result, we introduce the notion of {\em tolerance} ($\tau$). Also, as the accuracy of the model trained by decentralized learning increases, the space for model improvement is reduced. However, in such cases, incorporating learning contributions from encountered neighbors can still help the learners' model become more generalizable. 
Based on these findings, an updated learned model is considered valid if its resulting accuracy is either better than the input model's original accuracy or not worse than the input model by $\tau$. 
While we transmit vote results
as clear texts in our prototype,
existing research shows that voting can be secured to avoid bias in the result \cite{8887296}.

\begin{figure}[!t]
  \centering
  \includegraphics[width=\columnwidth]{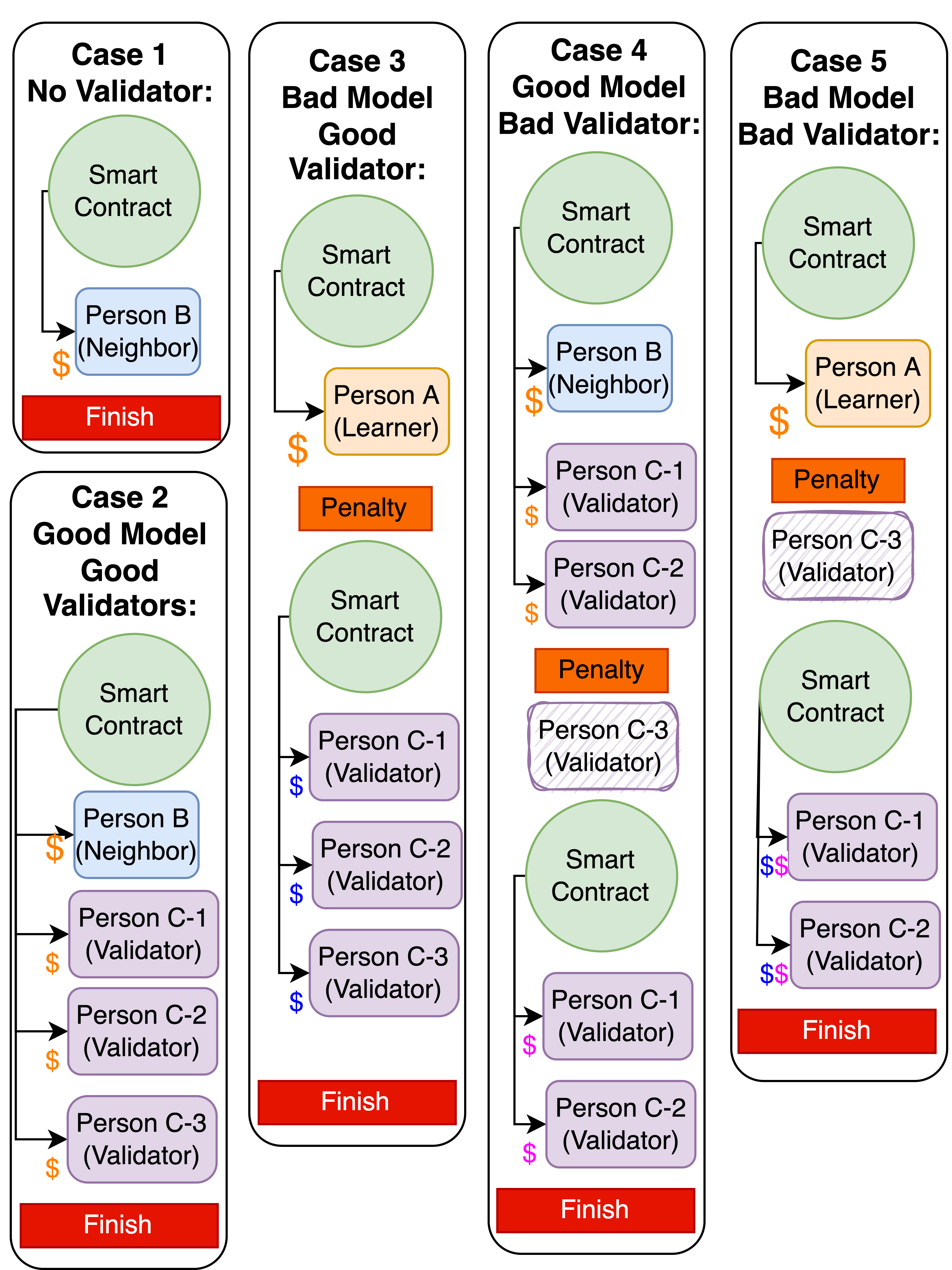}
  \caption{Finalizing the encounter ({\bf Steps 6-8)}.}
  \label{fig:step567} 
  \vspace{-.5cm}
\end{figure}

\begin{table}[]
\centering
\begin{tabular}{|c|p{7.5cm}|}
\hline
             & {\bf Explanation }                                                                                 \\ \hline\hline
$r$                 & amount of reward the learner requested to pay.                                          \\ \hline
$r_n$                 & amount of reward the neighbor receives.                                          \\ \hline
$r_v^t$                 & amount of reward for validators who voted yes.                                           \\ \hline
$r_v^f$                 & amount of reward for validators who voted no.                                           \\ \hline
$p_r^n$             & percentage of reward  the neighbor should receive.                              \\ \hline
$p_r^v $ &  percentage of reward validators should receive ($p_r^v = (1-p_r^n)$).                         \\ \hline
$n_v^t$             &  number of the validators who are voted yes.          \\ \hline
$n_v^f$             &  number of the validators that are voted no. \\ \hline
$V$                 & voting threshold; number of votes required to finalize                        \\ \hline
$\tau$              & Model accuracy tolerance.                                                               \\ \hline
\end{tabular}
\end{table}

\textbf{Step 6 through 8 (Finalization):} At the last
steps, the voting result can be computed if one of the following two conditions is satisfied: 1) there are more than $V$ (the voting threshold) votes or 2) the max voting time, which can be predetermined on smart contract, has passed. $p_r^n$ is the percentage of the award that will be given to the neighbor and $p_r^v$ is the percentage of the award that will be distributed among all of
the validators. Both $p_r^n$ and $p_r^v$ are
configurable per decentralized learning algorithm in the smart contract, and $p_r^v + p_r^n = 1$. We use $n_v^t$ to represent the number of validators that are voted the model has a reasonable contribution, and $n_v^f$ is the number voted oppositely. Finalization can be requested by any participant in the system and has the following possible outcomes, which is also summarized in Fig.~\ref{fig:step567}. Please note, in the figure, the brown dollar sign means the reward distributed from the {\bf Learner}, the blue dollar sign means the penalty from the neighbor, and the pink dollar sign means the penalty from the validator(s). For simplicity, in the figure we refer validators C-1 and C-2 as good ones while C-3 as the bad one in Cases 4 and 5.

\begin{description}
    \item \textbf{Case 1:} If there is no voting in the system ($n_v^t = n_v^f = 0$), we assume the model provided a solid contribution. 
        \begin{equation}
            r_n = r 
            \label{equ:1}
        \end{equation}
    \item \textbf{Case 2:} If all validators believe the {\bf Neighbor} has provided a reasonable contribution (i.e., $n_v^t \neq 0; n_v^f = 0$), the reward will be given as:
    \begin{equation}
        r_n = p_r^n * r 
        \quad  \text{and} \quad 
        r_v^t = \frac{p_r^v * r}{n_v^t}
        \label{equ:2}
    \end{equation}
    
    \item \textbf{Case 3:} If all validators believe the {\bf Neighbor} has not provided a reasonable contribution (i.e., $n_v^t = 0; n_v^f \neq 0$), the prepayment will be returned to the {\bf Learner} and the reward/penalty will be given as:
    \begin{equation}
        r_n = -r 
        \quad  \text{and} \quad 
        r_v^f = \frac{r}{n_v^t}
        \label{equ:3}
    \end{equation}
    
    \item[\textbf{Case 4:}] If a majority of validators believe the {\bf Neighbor} has provided a reasonable contribution but some others do not (i.e., $n_v^t \geq n_v^f >0$). The reward/penalty will be given as:
    \begin{equation}
        \begin{aligned}
            r_n = p_r^n * r 
            \quad  \text{and} \quad 
            & r_v^t = \frac{p_r^v * r - n_v^f *  r_v^f}{n_v^t} \\
           & r_v^f = - \frac{p_r^v * r}{n_v^t+n_v^f}
        \end{aligned}
        \label{equ:4}
    \end{equation}
    
    \item {\bf Case 5:} If a majority of validators believe the {\bf Neighbor}'s contribution is \textbf{not} a reasonable improvement but some others believe it is (i.e., $0<n_v^t < n_v^f$). The prepayment will be returned to the {\bf Learner} and the reward/penalty is:
    \begin{equation}
        \begin{aligned}
            r_n = -r  
            \quad  \text{and} \quad 
            & r_v^t =- \frac{r}{n_v^t+n_v^f}\\
           & r_v^f =  \frac{r - n_v^t * r_v^t}{n_v^f} 
        \end{aligned}
        \label{equ:5}
    \end{equation}
\end{description}

\subsection{Malicious Model Prevention}

As discussed previously, there are two significant possible classes of attacks on a system like iDML: incentive mechanism attacks and Learning attacks. Our approach is designed to prevent both classes.

{\bf Incentive mechanism attacks. }
The incentive mechanism attacks are those exposed by the incentive mechanism itself. In these situations, attackers aim to gain the reward associated with the learning process without actually engaging in meaningful collaboration. A setting where all participants of the system train and share models with other participants without ill intention is not realistic given participants may sabotage training by returning random model weights; such behavior is driven by the fact that model training at a single participant level is neither optimal nor efficient. Unfortunately, existing systems are not equipped with ways to detect such malicious behavior, which results in diminished accuracy and usefulnesss of the entire learning system. Such an attack can be addressed 
by leveraging iDML's validation steps (Steps 5 and 6 in Fig.~\ref{fig:system_layer}). 

Furthermore, the validation result may lead to penalties for the attacker. Step 0 in Fig.~\ref{fig:step0} asks the neighbor to stake their token before collaboration, which provides a mechanism for the smart contract to penalize a dishonest {\bf Neighbor} participant. In Eq.~\ref{equ:3} and Eq.~\ref{equ:5}, we guarantee that the penalty the attacker may receive is equal to or slightly larger than what they may earn (e.g., when the {\bf Neighbor} is also a ``positive" validator), which makes the mathematical expectation for the incentive mechanism attack equal to or less than 0.

{\bf Learning attacks.}
Learning attacks seek to cause the output of the learning process to misclassify an input, which is also commonly referred to as a poisoning attack. In iDML, the staking and the validation steps are also effective at preventing such learning attacks. In the case that {\bf Neighbor} participants are dishonest in any of their actions, a penalty will be applied to the token they staked to the smart contract.
\section{Evaluations}
\label{evaluation}
In this section, we evaluate iDML with two 
state-of-the-art
decentralized learning algorithms: gossip learning~\cite{colin2016gossip} and opportunistic collaborative learning (OppCL)~\cite{9439130}.

\subsection{Datasets and Models}
Our evaluation relies on two widely used classification datasets: CIFAR-10 \cite{cifar_10}, which consists of 60,000 32*32 colorized images in 10 classes, and Fashion-MNIST \cite{Fashion-MNIST}, which includes 70,000 28x28 grayscale images in 10 classes. We use 10,000 images in each dataset as the testing dataset. For the training data, we split the remaining samples among the potential training devices (i.e., the {\bf Neighbor} participants) in iDML using two methods: (1) Independent and Identical Distribution (IID): the training data is evenly distributed, without overlap, to all of the simulated participants; (2) Non-Independent and Identical Distribution (Non-IID): each {\bf Neighbor} participant has the same amount of data, without overlap, but each neighbor has only 2 classes of the data.

For both learning algorithms, we choose those learning models that are 
small enough to be compatible with training on resource-constrained devices and adaptle to different model sizes. For the gossip learning approach that we employ~\cite{colin2016gossip}, all participants rely on the same model structure; upon an encounter, the devices simply exchange snapshots of their current model parameters, which are aggregated locally before training continues.  We use 10 convolutional layers, max pooling, and dropout layers. In contrast, for our opportunistic learning model, we use three different sizes for the convolutional neural network (CNN) models, which have 6, 10, ad 14 layers, max pooling, and dropout layers.

Decentralized and opportunistic learning relies on encounters between participating nodes. We rely on commonly used datasets of contact patterns  \cite{ozella2021using,upb-mobility2011-20120618, upb-hyccups-20161017}. 
However, these datasets are biased towards some popular nodes which are not suitable for our experiments.
Instead of using these encounter patterns directly, we use insights from these datasets
to randomly generate larger encounter patterns that provide evenly distributed contact patterns for 50 simulated participants.

\subsection{Evaluation Platform and Setting}
We implement our smart contract in Solidity\footnote{\url{https://soliditylang.org/}} and deploy it on Ganache\footnote{\url{https://trufflesuite.com/ganache/}}, a smart contract tool. The decentralized opportunistic learning simulation system is implemented using Python with TensorFlow. The simulation is running on a Workstation with Ubuntu 18.04 LTS System, i7-9800X CPU, 128G Memory, and NVIDIA GeForce RTX 2080 Ti GPU.

iDML's validation process depends on the decentralized learning algorithm. Thus, we separately implemented the validation process for each algorithm. For gossip learning, we let the {\bf Learner} keep a copy of the model before performing the receive-merge-train cycle, and share both the old and new model with the {\bf Validator}. For OppCL, the {\bf Learner} will not merge the model after it has been received
but
share both the model with the {\bf Validator} to validate. For both algorithms, the {\bf Validator} will evaluate the accuracy of the model and vote based on the result. The other parameters used in our experiments are listed in Table~\ref{tab:parameters}.

\begin{table}[]
\centering
\caption{Simulation Parameters}
\label{tab:parameters}
\begin{tabular}{|p{7cm}|l|}
\hline
{\bf Parameter}              & {\bf Value}               \\ \hline\hline
Stake threshold for a user to become {\bf Neighbor} or {\bf Validator} & 100   \\ \hline
Voting threshold $V$ & 3   \\\hline
Max Voting waiting time (block)  & 50000 \\ \hline
Number of tokens each participant exchanged  & 1000000   \\ \hline
Number of tokens each {\bf Neighbor}/{\bf Validator} staked initially & 200   \\ \hline
Tolerance $\tau$ for F-MNIST $\|$ IID $\|$ OppCL (For Fig.~\ref{fig:attack_acc} and \ref{fig:blockchain_result})   & 0.030 \\ 
\hline
\end{tabular}
\end{table}

\subsection{Experimental Results}

\begin{figure}[!t]
 \centering
 \includegraphics[width=0.48\textwidth]{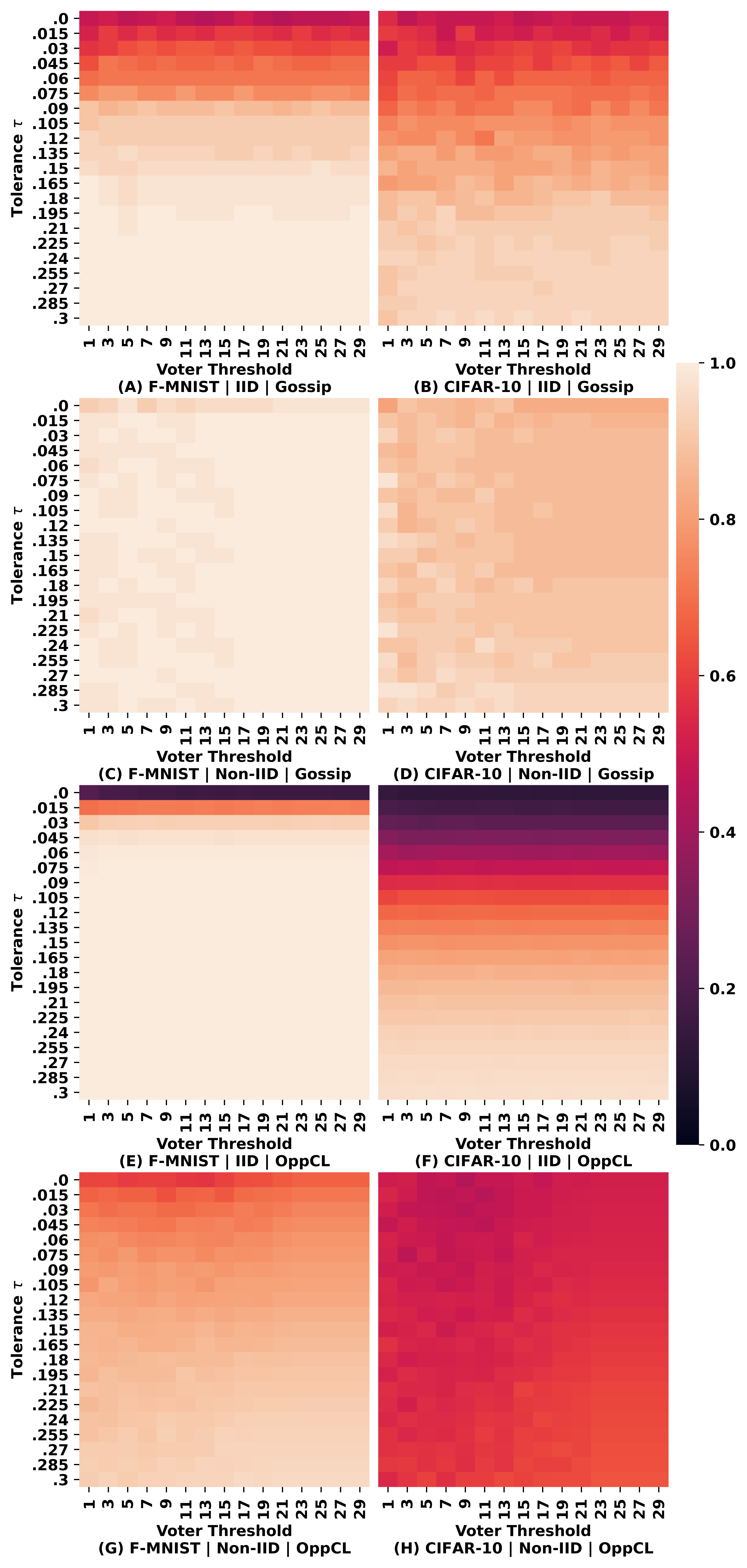}
 \caption{Voting accuracy with different $\tau$ and voters settings}
 \label{fig:tol_val}
\end{figure}
In our first and primary experiment, our goal is to evaluate the effectiveness of iDML in providing incentives to participate in decentralized opportunistic learning. In particular, we evaluated the 
voting accuracy under the settings of different number of validators and training tolerance ($\tau$) for both gossip learning and OppCL. 
Note, the tolerance value gives a lower-bound of the trained model's accuracy ($\tau$ definition is on Step 5). The voting accuracy reflects how many training instances are validated by our validation algorithm. The instance is regarded as validated if the model's accuracy tested on the validation set satisfies the tolerance setting. If the model is more generalizable, it is more likely the model can pass our validation.
 
In particular, Fig.~\ref{fig:tol_val} shows iDML's
effectiveness in verifying whether a {\bf Neighbor} created a valid contribution for the {\bf Learner}, given a specific threshold and number of {\bf Validator}s (i.e., voters).
Fig.~\ref{fig:tol_val} shows in IID settings, both learning algorithms show stable voting accuracy pattern across different number of voters. Through another in-depth study using very few number of voters (skipped due to brevity), we observe three voters is a good choice in the settings. Though in non-IID settings,  
the voting accuracy
fluctuates a bit among different number of voters, considering the costs of having additional voters, smaller number of voters are desirable (e.g., 3).

We also observe the different behavior of Gossip and OppCL in IID and Non-IID settings.  It shows our adapters for both learning algorithms work as the trend shows the native characteristics of both algorithms. 
For instance, Gossip learning uses a {\em merge-update-send} cycle and OppCL adopts a {\em share-train-return} cycle. The difference (Gossip is more generalizable than OppCL per training instance) caused more variance in voting accuracy in the models with OppCL, compare with the Gossip learning.

One practical value of iDML is that, under the assumption that participants’ behaviors will be driven by clearly-defined incentives, the explicit incentive mechanism we defined will result in the effective engagement of system participants and a higher resulting model accuracy from collaboration. Multiple rounds of simulations show that the current system design awards laborious {\bf Neighbor} participants, while those who 
hinder training by any means are marginalized from participation in the system.
In real-world applications, indeed, human behaviors may be more complex and may not be as directly driven as machines are. Yet, many applied cases show that a clearly defined reward (e.g., monetary value and promotions) encourages desired behaviors.

In our final experiment, we assess the degree to which iDML also prevents learning attacks. We assign different numbers of learning attackers. Each of these attackers flip the label of 
images
in their dataset (e.g. they, mark trucks as airplanes and/or deer as dogs). We inject these attackers into the system randomly and check the model accuracy after every encounter round for all of the participants with the testing dataset. 

\begin{figure}
 \centering
 \includegraphics[width=0.48\textwidth]{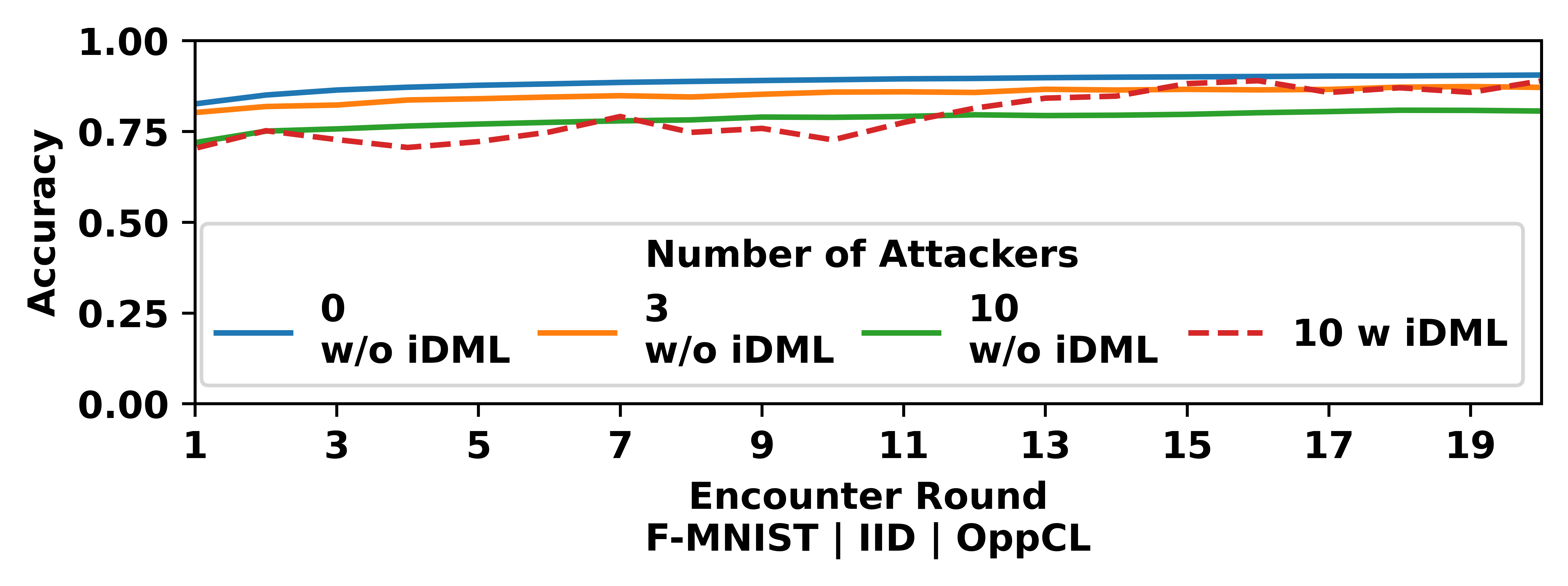}
 \caption{Model accuracy for different numbers of attackers}
 \label{fig:attack_acc}
\end{figure}

\begin{figure}
 \centering
 \includegraphics[width=0.48\textwidth]{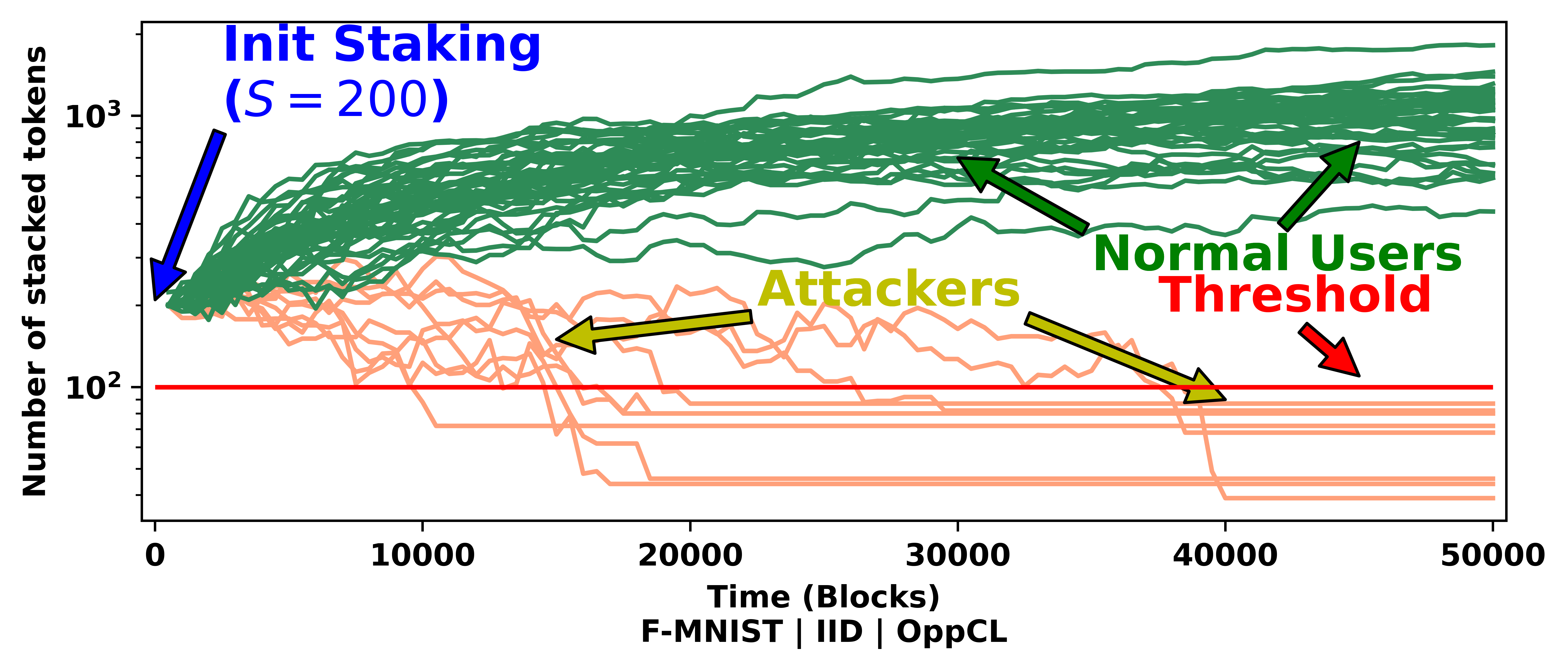}
 \caption{Progression of staked tokens for participants during the training process}
 \label{fig:blockchain_result}
\end{figure}

As a baseline, we first experiment
without iDML.
We plant $0$,$3$, and $10$ attackers in the system respectively.
Fig.~\ref{fig:attack_acc} shows that under OppCL, an increased number of attackers causes the system to incur a delay in accuracy increase. Further, the system ceases to increase learning accuracy after several rounds of encounters. The more attackers in the system, the lower the ending model's average accuracy.
Then, we redo the experiment with $10$ attackers with iDML;
the result is shown as the red dotted line in Fig.~\ref{fig:attack_acc}. It shows that iDML first follows a similar trend of the $10$ attackers experiment and then 
recovers.
To understand the cause of this behavior, we examine the amount of staking in the smart contract for each encounter around with all of the participants (Fig.~\ref{fig:blockchain_result}). The results using different algorithm and dataset combinations are similar to Fig.~\ref{fig:attack_acc} and Fig.~\ref{fig:blockchain_result} and are therefore omitted for brevity.

We believe that at the first stage (before $40000$ blocks), even though iDML rejected some of the attacker's models, these attackers are still active in the learning system. In the meantime, iDML continues to penalize those attackers as they provide malicious models and/or act as malicious ``positive" validator. After the turning point ($40000$ blocks), the staking for attackers falls below the threshold, and they can no longer collaborate in the system. After that, iDML recovers the learning system from the attacks and the model accuracy begins improving once again.

\subsection{Discussion}
While these results show the promise of an approach like iDML, there remain some additional constraints and limitations associated both with its reliance on blockchain and the design of the system itself.

\begin{table*}[!t]
\centering
\caption{Example cost of one learning encounter in iDML(cost baselines from Jan 17, 2023 04:27 UTC)}
\label{tab:gas}
\begin{tabular}{|l|c|c|c|}
\hline
{\bf Step}  & {\bf Gas Used} & {\bf Cost (ETH Network)} & {\bf Cost (Polygon Network)} \\ \hline
2. Pre-Training Check (Learner) & 43919 & \$0.9623 & \$0.0025 \\ \hline
4. Learning Complete (Neighbor) & 96172 & \$2.1072 & \$0.0055 \\ \hline
4. Learning Complete (Learner) & 46921 & \$1.0281 & \$0.0027  \\ \hline
6. Votes (Validator) & 78869 & \$1.7281 & \$0.0045 \\ \hline
7. Result Check  and 8. Distribute Award (Any one party) & 191344  & \$4.1925 & \$0.0109 \\ \hline
{\bf Total} & 457225  & \$10.0182 & \$0.0261 \\ \hline
\end{tabular}
\end{table*}

\subsubsection{Blockchain}
The proposed system is a decentralized platform that utilizes the blockchain network to provide secure and efficient data storage and sharing. However, there are certain limitations that come with using blockchain technology.
\begin{itemize}
    \item \textbf{Operation cost}: 
    Most operations in a block chaian network (e.g., (EVM opcodes), including stack (POP, PUSHX, DUPX, SWAPX), memory (CALLDATACOPY, CODECOPY, etc.), and storage (SLOAD and SSTORE)\cite{EVM},)
    incur a cost known as ``gas''; the used gas is then converted to a monetary fee that is dependent on the cost baseline of the network being used. Table~\ref{tab:gas} shows the gas used on two example networks for a single encounter in iDML. In employing iDML, the application designer must navigate the tradeoff between the benefit to deploying incentives against their cost---iDML will not be appropriate for all applications, but Table~\ref{tab:gas} shows that the costs (1)~can be made reasonable and (2)~can be estimated in advance. In the near future, the cost can be lower with new technologies, such as Polygon Miden \footnote{\url{https://polygon.technology/solutions/polygon-miden}}.
    
    \item \textbf{Security risk:} As the proposed system relies on the safety and reliability of the blockchain, any security issues that occur on the blockchain may also affect the proposed system. For example, if a hacker successfully performs a $51\%$ attack and gains control of the blockchain, they could potentially tamper with the data stored on the proposed system \cite{attack}.

    \item \textbf{Limited privacy}: If the smart contract is deployed to a public chain, such as ETH or Polygon, the information can be accessed by the public, which may be a concern for leaking sensitive data such as personal encounter information. Therefore in the design of iDML, we intentionally keep model weights and data distributions off the blockchain to reduce the privacy risk.
\end{itemize}

\subsubsection{System}
The system relies on validators to determine whether a neighbor's contribution is valid, but due to the nature of the machine learning task and the randomness of datasets and the learning process, fluctuations in accuracy are natural and not necessarily indicative of an invalid contribution. Depending on the particular validator(s) a learner encounters, the validation result may vary. We have applied a validation tolerance value ($\tau$) to reduce this issue, but it cannot be completely eliminated. Additionally, the accuracy may not always fully reflect the improvement of a model. Other evaluation metrics, such as recall, precision, F1, and loss, may better reflect the improvements. Future work could explore choosing  performance metrics that better fit a specific learning task, and iDML could be updated to allow different learners to  mandate the use of different metrics. In this study, however, we rely on accuracy to show the potential for iDML to incentivize neighbor participation. 

Finally, iDML has been specifically designed as an incentive mechanism for decentralized machine learning. We have included components of the design to prevent or mitigate potential attacks on the system, but these are not comprehensive. For instance, iDML does, by design, mitigate (though it does not completely prevent) model poisoning attacks that may be launched by malicious neighbors by relying on the validators to sign on each neighbor's contribution. On the other hand, if a group of attackers conspires to pose as malicious neighbors and validators, the system cannot mitigate this attack. In the future, by evalauting iDML in the context of real encounter patterns in mobile networks, we can test the difficulty, cost, and potential damage associated with launching  such an attack
As a more nuanced example, the tolerance $\tau$ is useful because it allows for some expected fluctuations in model accuracy, but a persistent attacker could attempt to manipulate a model by working within the tolerance range to shift the model's direction. However, to successfully launch such an attack, a neighbor would have to be able to influence a learner's model repeatedly over time (assuming $\tau$ is reasonably small), so an easy way to mitigate this attack is to ensure diversity in the selected neighbors.
\section{Conclusion and Future Work}
\label{conclusion}
We have presented a framework called iDML for incentivizing decentralized and opportunistic machine learning. The evaluation results show that the system can effectively compensate neighboring devices that provides communication and computational resources to assist other participants' in their model learning tasks. In addition, the results show the system can detect attackers in the system and avoid malicious models being merged into the user's model. Also, the evaluation of different decentralized learning algorithms with different data and data distributions shows the wide range of the use cases of our proposed iDML system.
In future work, one can evaluate the system with large-scale real-world contract patterns. On the other side, to better incentivize participants, a data item's rarity could also be considered and provided corresponding rewards.

\bibliographystyle{IEEEtran.bst}
\bibliography{references.bib}

\end{document}